\documentclass[runningheads]{llncs}
\usepackage[T1]{fontenc}
\usepackage{graphicx,verbatim}
\usepackage{amssymb}
\usepackage{amsmath}
\usepackage{algorithm}
\usepackage{algpseudocode}
\usepackage{hyperref}
\usepackage{cite}
\usepackage{graphicx}
\usepackage{wasysym}
\usepackage{marvosym}
\usepackage{multirow}
\usepackage{adjustbox}
\usepackage{orcidlink}

\begin{document}
\title{HRVVS: A High-resolution Video Vasculature Segmentation Network via Hierarchical Autoregressive Residual Priors}
\titlerunning{HRVVS: A High-resolution Video Vasculature Segmentation Network}
%
%
\author{
Xincheng Yao\inst{1}
\and
Yijun Yang\inst{1}$^\dagger$
\and
Kangwei Guo\inst{2}
\and
Ruiqiang Xiao\inst{1}
\and
Haipeng Zhou\inst{1}
\and
Haisu Tao\inst{2}
\and
Jian Yang\inst{2}
\and
Lei Zhu\inst{1,3} $^\textrm{\Letter}$
}
\authorrunning{F. Author et al.}
%
\institute{The Hong Kong University of Science and Technology (Guangzhou), China 
 \and 
Southern Medical University, Guangzhou, China
\and
The Hong Kong University of Science and Technology, Hong Kong, China\\
}
\maketitle              
\renewcommand\thefootnote{}\footnotetext{$^\dagger$ Project Lead.}
\begin{abstract}
The segmentation of the hepatic vasculature in surgical videos holds substantial clinical significance in the context of hepatectomy procedures. However, owing to the dearth of an appropriate dataset and the inherently complex task characteristics, few researches have been reported in this domain. To address this issue, we first introduce a high quality frame-by-frame annotated hepatic vasculature dataset containing 35 long hepatectomy videos and 11442 high-resolution frames. On this basis, we propose a novel high-resolution video vasculature segmentation network, dubbed as HRVVS. We innovatively embed a pretrained visual autoregressive modeling (VAR) model into different layers of the hierarchical encoder as prior information to reduce the information degradation generated during the downsampling process. In addition, we designed a dynamic memory decoder on a multi-view segmentation network to minimize the transmission of redundant information while preserving more details between frames. Extensive experiments on surgical video datasets demonstrate that our proposed HRVVS significantly outperforms the state-of-the-art methods. The source code and dataset will be publicly available at \href{https://github.com/scott-yjyang/HRVVS}{https://github.com/scott-yjyang/HRVVS}.

\end{abstract}
\section{Introduction}
Hepatectomy is a set of surgical procedures for local liver lesions, such as liver tumors, liver injuries, liver abscesses, and etc.. Given the rich blood supply in the liver, effective control of bleeding during surgery is pivotal for the success of liver resection~\cite{smyrniotis2005vascular,ali2025objective}. Specifically, during the operation, surgeons need to focus on two types of blood vessels, the Glisson sheath and the hepatic vein. The segmentation of these hepatic vasculature in surgical videos can provide precise positioning for surgeons to prevent surgical bleeding by hemoclips during hepatectomy, which makes it has great important clinical significance. Nevertheless, the fat and muscle around the vasculature generate significant redundant in formation, makinges the model difficult difficult to segment the correct tissue in the video. Previous works~\cite{guo2020novel, yang2021hcdg,yang2023mammodg, li2024automated, shi2024centerline} in hepatic vascular segmentation mainly focused on medical images from CT or MRA before surgery. However, these methods can not directly pinpoint the location of vasculature during the actual surgical procedure. Therefore, we are considering building a high-resolution video segmentation network to address this issue. Through collaboration with the hospital, we collect a dataset containing 35 long high-resolution videos with a total of 11442 frames, which provided the foundation for our model training. To the best of our knowledge, ours is the first work dedicated to this task.

\begin{figure}[t]
    \centering
    \includegraphics[width=\linewidth]{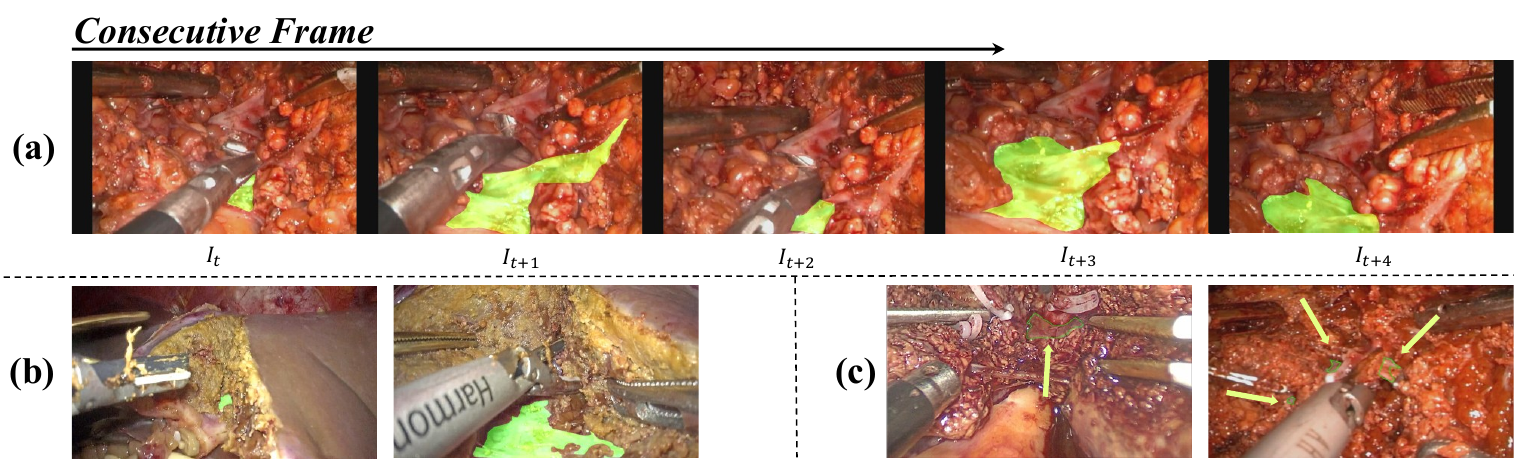}
    \caption{Main challenges in hepatic vasculature segmentation. Fluorescent green shadow is used to show the location of vasculature in \textbf{(a)} and \textbf{(b)}, the arrow and green outline are used to point to the corresponding location in \textbf{(c)}. \textbf{(a)} Discontinuities between frames and abrupt positional transformations. \textbf{(b)} Significant variations of vessels in different contexts. \textbf{(c)} Similarities in the outline of vessels and surrounding tissue and segments by the surrounding tissue.}
    \label{fig: challenges}
\end{figure}

In recent years, numerous video segmentation methods have emerged for medical imaging, exemplified by approaches such as Vivim~\cite{yang2024vivim,yang2025vivim}, which utilizes a state space model, and Ji \textit{et al.} introduce SUN-SEG dataset for polyp segmentation in colonoscopy videos, alongside the PNS+ algorithm \cite{ji2022video}. Furthermore, the advent of SAM2 \cite{ravi2024sam} has inspired a series of video segmentation techniques \cite{liu2024surgical, zhu2024medical, mansoori2024polyp}, demonstrating remarkable efficacy in medical video segmentation. However, these methods and their associated datasets are not optimized for high-resolution tasks, and their performance is often compromised in complex surgical scenarios. In our dataset, the segmentation of hepatic vasculature presents specific challenges, as illustrated in Fig \ref{fig: challenges}. These include \textit{frame discontinuities} and \textit{abrupt positional changes} (Fig \ref{fig: challenges} (a)), \textit{significant variations in vessel appearance} due to differing anatomical contexts and imaging conditions (Fig \ref{fig: challenges} (b)), and the \textit{similarity between vessel outlines and surrounding tissue}, which can lead to segmentation errors (Fig \ref{fig: challenges} (c)). These factors complicate the task of maintaining segmentation continuity and accuracy across frames.

To address these issues, our approach in designing a high-resolution surgical video segmentation model focuses on two critical aspects: preserving detailed features within the current frame and minimizing computational load and cumulative errors from redundant frame-to-frame propagation. For the former, we employ a pretrained VAR model \cite{tian2024visual} as residual priors to enhance the hierarchical encoder framework, refining it through adapter-based training. For the latter, we introduce a dynamic memory decoder featuring a Multi-view Spatiotemporal Interaction Module (MSIM) and a Dynamically Weighted Fusion Module (DWFM) for memory propagation. Our method demonstrates state-of-the-art performance when benchmarked against the latest segmentation approaches, effectively overcoming the identified challenges associated with high-resolution tasks in complex surgical environments.

In conclusion, our contributions are: (1) \textbf{We develop a high-resolution video segmentation model for hepatic vasculature}, demonstrating the effectiveness of VAR in segmentation tasks. (2) \textbf{We introduce the first high-resolution video hepatic vasculature segmentation dataset under surgical scene}, which can be seen as a benchmark dataset for a completely new task. (3) Extensive experiments have been conducted on our dataset, \textbf{demonstrating the superiority of our proposed method}.

\section{Method}
\begin{figure}[t]
    \centering
    \includegraphics[width=\linewidth]{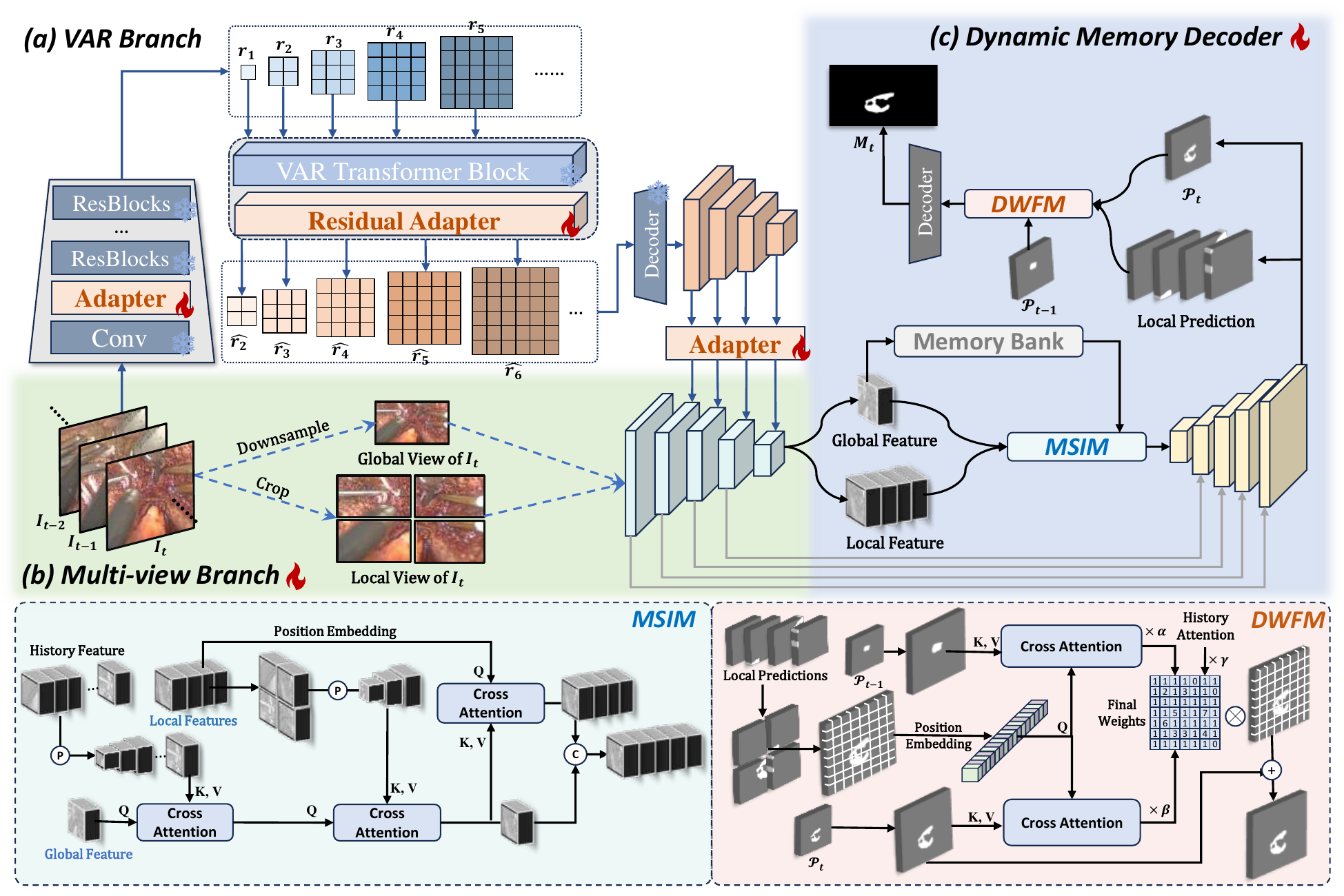}
    \caption{Pipeline of the proposed HRVVS. Our method comprises three main components: (a) ``VAR Branch'', a multi-scale generation branch based on visual autoregressive modeling equipped with adapters. (b) ``Multi-view Branch'' is based on a hierarchical encoder with five different views of the current frame. (c) the ``Dynamic Memory Decoder'' is a decoder of our network, which includes a multi-scale decoder, a memory bank, a Multi-view Spatiotemporal Interaction Module (MSIM), and a Dynamic Weights Fusion Module (DWFM). Below the pipeline we show the detailed structure of the MSIM and DWFM.}
    \label{fig: framework}
\end{figure}
\subsection{Overview}
Fig \ref{fig: framework} shows the overall framework of the proposed segmentation model. For a high-resolution frame input, we extract its multi-level features by a dual-branch encoder based on VAR and Swin Transformer~\cite{liu2021swin}. To tackle the aforementioned challenges (Fig~\ref{fig: challenges}), we proposes a memory-augmented decoder that integrates long short-term memory architecture, comprising a Multi-view Spatiotemporal Interaction Module (MSIM) and a Dynamic Weights Fusion Module (DWFM). MSIM preliminarily updates the local, global, and historical features through multi-dimensional spatiotemporal feature interaction mechanisms. The updated local and global features from MSIM will be sent into the multi-level decoder, which will also have the residual input from the corresponding layers of the multi-view encoder.
After we get the local and global features before the last layer of decoder, they will be fused together with the global feature of the previous frame from the memory bank as a reference of weights in DWFM.
The final prediction will be obtained from the fused feature of DWFM. The details of each module are described in the following subsections.

\subsection{Dual-branch Residual Prior Encoder}
Visual auto-regressive modeling (VAR)~\cite{tian2024visual} is renowned for its scalable auto-regressive generation capability. Its multi-scale unified quantization enables consistent image representation across different scales, effectively capturing both global context and fine-grained details. By extracting hierarchical features from the VAR branch and incorporating them as residual priors into the downsampling layers of the multi-view branch, VAR enhances the information flow within the multi-view branch, improving feature representation.

Given a specific frame \(I_t \in \mathbb{R}^{3\times H\times W}\) in an HR video \(\mathcal{V} = \{I_i \mid i = 1,2,\cdots,n\}\), we extract its hierarchical features separately through the multi-view branch. For VAR branch, we got the multi-scale consistent features from the VAE decoder and add them through the leanrable project layers as residual priors to the downsampling layers of multi-view branch. In the multi-view branch, we process high-resolution images by performing center quartering operations and downsampling on \(I_t\) respectively to obtain the local view $\{L_m\}_{m = 1}^4 \in \mathbb{R}^{3\times h\times w}$ and global view \(G \in \mathbb{R}^{3\times h\times w}\) of \(I_t\), respectively, where $(H, W) = (2h, 2w)$. 
The features derived from the encoder are denoted as $\{\mathcal{F}^i_m\}_{m = 1}^5 = \{{\{\mathcal{L}^i_m\}_{m=1}^4}, \mathcal{G}^i\}$, where $i \in [1, 5]$ represents features extracted from the i-th layer of encoder. In the VAR branch, by using the pre-trained VQ-VAE and VAR weights, we freeze the backbone network, and finetune adapters incorporated between the VAE encoder and the VAR transformer blocks. 
As shown in Eq.~\ref{eq:var_next_scale}, $\mathcal{A}$ and $\mathcal{T}$ represent the VAR adapter and VAR transformer block respectively, respectively, while $\theta_{\mathcal{A}}$ represents the learnable parameters in $\mathcal{A}$:
\begin{equation}
r_{k} = \mathcal{A}\left(\mathcal{T}([s], r_{1}, r_{2}, \cdots, r_{k - 1}); \theta_{\mathcal{A}}\right)
\label{eq:var_next_scale}.
\end{equation}
Another adapter is adopted to project VAE features into multi-view latent space. 
We denote the fused features as ${\mathcal{F'}^i} \in \{\mathcal{L'}^i, \mathcal{G'}^i\}$ and store $\mathcal{G'}^i$ in the memory bank as the current frame global feature.

\subsection{Multi-view Spatiotemporal Interaction Module}
To update the multi-view features and the global features from the memory bank, we have designed a Multi-view Spatiotemporal Interaction Module (MSIM).
Inspired by \cite{yu2024multi,yang2023diffmic,yang2025diffmic}, we update the local and global features by a multi-head cross-attention (MHCA).
With respect to historical frame features, given the rapid dynamics of surgical scene changes during the procedure, we contend that their informational relevance increases with temporal progression. To mitigate the computational complexity inherent in attention mechanisms, we employ exponential downsampling rates for historical frames in the memory bank, with the rate being lower for earlier historical frames and append positional encodings to each frame to facilitate efficient querying.
Before the multi-view interaction process, we introduce the multi-scale memory features as the reference to update the global features. 

As shown in Fig \ref{fig: framework}, during the forward propagation, $\{\mathcal{L'}^5_m\}_{m=1}^4$ are first rearranged and position-encoded. 
When the number of historical frames reaches the upper bound, the historical frames are downsampled and position-encoded to generate multi-scale memory concatenated tokens $\mathcal{H}_n$. MHCA is then employed to compute the attention for the historical frames, thereby updating the global features $\mathcal{G}^5$ as Eq. \ref{eq: hitory_fuse}, where \( Q \) are the tokens from $\mathcal{G'}^5$, and \( K \) and \( V \) are $\mathcal{H}_n$:
\begin{equation}
    \mathcal{G}_h = \mathcal{G'}^5 + \text{Dropout}\left(\text{MHCA}\left(Q, K, V\right)\right)
    \label{eq: hitory_fuse},
\end{equation}
In the other branch, the local features of the current frame are pooled to generate features at different scales. We also use MHCA to further update the global feature \(\mathcal{G}_h\) and get $\mathcal{G}_{msim}$, which is updated by $\{\mathcal{L'}^5_m\}_{m=1}^4$.
The updated local features $\{\mathcal{L}_{mism}\}_{m=1}^4$ are then rearranged and concatenated with the updated global features to be the input of the decoder:
\begin{equation}
    \{\mathcal{L}_{mism}\}_{m=1}^4 = \{\mathcal{L}^5_m\}_{m=1}^4 + \text{Dropout}\left(\text{MHCA}\left(Q_m, K + p_{\text{poses}}, V\right)\right)
    \label{eq: lmism}.
\end{equation}

\subsection{Dynamic Weights Fusion Module}
The weights for fusing high-resolution sub-images critically determine the quality of the final generated image~\cite{yang2023video,yang2024genuine,liang2025global}. To avoid boundary distortion and information inconsistency in fusion process, we proposed DWFM.
Unlike MSIM, we aim to optimize the multi-view fusion process via weight variations. Specifically, we further divide the 4 local features into $4\times16$ small patches, and assign corresponding weights to each patch to reduce the boundary fragmentation caused by local attempts to directly aggregate. We take each local patch as $Q$ and calculate MHCA with the current global feature and the last global feature separately, obtaining the corresponding importance weights $\mathcal{W}_{final}^i$.

Excessive focus on the global features of the current frame can lead to a loss of reference to changes in adjacent frames. We compute Weights A and Weights B from the current global feature $\mathcal{P}_t$ and the previous global feature $\mathcal{P}_{t-1}$, labeling them as $\mathcal{W}_g^t$ and $\mathcal{W}_l^t$ for the t-th frame, respectively. For the first frame in each video, only its global weight $\mathcal{W}_g^0$ is calculated, and it is stored as the historical weight $\mathcal{W}_h^1$ for the next frame's prediction. For each subsequent frame's patches, the weight score of each patch is continuously updated based on the historical weight $\mathcal{W}_h^t$, the current global feature weight $\mathcal{W}_g^t$, and the previous frame's weight $\mathcal{W}_l^t$. The expressions for updating historical weights and calculating current weights are shown in equations \ref{eq: dwfm equation} and \ref{eq: history weight update equation}, respectively.
\begin{equation}
\mathcal{W}_{final}^t = 
\begin{cases}
\mathcal{W}_g^0, & t = 0 \\
\alpha \times \mathcal{W}_l^t + \beta \times \mathcal{W}_g^t + \gamma \times \mathcal{W}_h^t, & t > 0
\end{cases}
\label{eq: dwfm equation},
\end{equation}
\begin{equation}
\mathcal{W}_h^{t+1} = \delta \times \mathcal{W}_h^t + (1 - \delta) \times \mathcal{W}_{final}^t
\label{eq: history weight update equation}.
\end{equation}
\section{Experiments}
\subsection{Dataset and Experimental Settings}
\subsubsection{Hepa-SEG Dataset.} 
We introduce \textbf{Hepa-SEG}, the first vasculature segmentation dataset for hepatectomy.
The dataset consists of 35 hepatectomy videos, totaling 11,442 frames with a resolution of 1080$\times$1920. Each video contains approximately 8 minutes of continuous frames from the liver transection stage, where every frame is manually annotated. The dataset includes two vasculature types: the Glisson sheath and the hepatic vein. The data is randomly split into training, validation, and test sets with a ratio of 7:1:2.


\subsubsection{Implementation Details.}
All experiments are conducted on a single NVIDIA A800 GPU. Our model is trained for 15 epochs with a batch size of 32. A sliding window sampler is used to ensure that each batch contains consecutive frames. We optimize the model using Adam with an initial learning rate of $1\times10^{-5}$, which is decayed using a polynomial scheduler with a decay rate of 0.9.


\begin{table}[t]
  \caption{Quantitative Comparison with Different Methods on Hepa-SEG. The best values are highlighted in bold. $\uparrow$ denotes that a higher score is better.}
  \centering
  \begin{adjustbox}{width=\textwidth}
  \begin{tabular}{c|c|c|c|c|c|c|c}
  \hline
  \textbf{Methods} & \textbf{Venue} & \textbf{Type} & \textbf{Jaccard \boldmath{$\uparrow$}} & \textbf{Dice \boldmath{$\uparrow$}} & \boldmath{${S}_\alpha \uparrow$} & \boldmath{$F_\beta^\omega \uparrow$} & \boldmath{$E_\phi^{mn} \uparrow$} \\
  \hline
  PraNet~\cite{fan2020pranet}    & \textit{$MICCAI_{20}$} & image      & 0.3569 & 0.4586 & 0.6875 & 0.5124 & 0.8135 \\
  LDNet~\cite{zhang2022lesion}   & \textit{$MICCAI_{22}$} & image      & 0.2322 & 0.2929 & \textbf{0.8355} & 0.2798 & 0.8331 \\
  ISNet~\cite{qin2022highly}    & \textit{$ECCV_{22}$} & image        & 0.1982 & 0.2576 & 0.7854 & 0.2710 & 0.8103 \\
  HitNet~\cite{hu2023high}       & \textit{$AAAI_{23}$} & image        & 0.4481 & 0.5700 & 0.4851 & 0.5434 & 0.8276 \\ 
  \hline
  SLT-Net~\cite{cheng2022implicit}   & \textit{$CVPR_{22}$} & video          & 0.2825 & 0.4904 & 0.6521 & 0.4097 & 0.6729 \\
  Vivim~\cite{yang2024vivim}      & \textit{$TCSVT_{24}$} & video      & 0.4480 & 0.5801 & 0.7511 & 0.5801 & 0.8380 \\
  Med-SAM2~\cite{zhu2024medical} & \textit{$Arxiv_{24}$} & video       & 0.3470 & 0.4555 & 0.6728 & 0.4552  & 0.5268 \\
  SALI~\cite{hu2024sali}     & \textit{$MICCAI_{24}$} & video          & 0.5239 & 0.6424 & 0.7748 & 0.6496 & 0.8405 \\
  MemSAM~\cite{deng2024memsam}      & \textit{$CVPR_{24}$} & video     & 0.1337 & 0.2126 & 0.4642 & 0.2369 & 0.4683 \\ 
  \hline
  \textbf{HRVVS(Ours)}   & -- -- & video & \textbf{0.5405} & \textbf{0.6532} & 0.7878 & \textbf{0.6769} & \textbf{0.8711} \\ 
  \hline
  \end{tabular}
  \label{tab:comparison_hepa}
  \end{adjustbox}
\end{table}

\subsection{Comparisons with State-of-the-Arts}
\subsubsection{Baselines and Metrics.} 
We evaluate HRVVS against nine state-of-the-art segmentation methods on the \textbf{Hepa-SEG dataset}, including four image-level and five video-level approaches. Specifically, the baselines comprise two high-resolution segmentation methods (i.e., HitNet\cite{hu2023high} and ISNet\cite{zhang2022isnet}), six medical image segmentation methods (i.e., PraNet\cite{fan2020pranet}, LDNet\cite{zhang2022lesion}, Vivim\cite{yang2024vivim}, Med-SAM2\cite{zhu2024medical}, SALI\cite{hu2024sali}, and MemSAM\cite{deng2024memsam}), and one general segmentation method (i.e., SLT-Net\cite{feng2022slt}). 
For quantitative evaluation, we adopt five commonly used metrics~\cite{lu2024diff}: Jaccard index, Dice coefficient, Structure-measure ($\mathcal{S}_\alpha$)\cite{fan2017structure}, F-measure ($F_\beta^\omega$)\cite{achanta2009frequency}, and Enhanced-alignment measure ($E_\phi^{mn}$)\cite{fan2021cognitive}.

\subsubsection{Quantitative Comparison.} 
\begin{figure}[t]
    \centering
    \includegraphics[width=\linewidth]{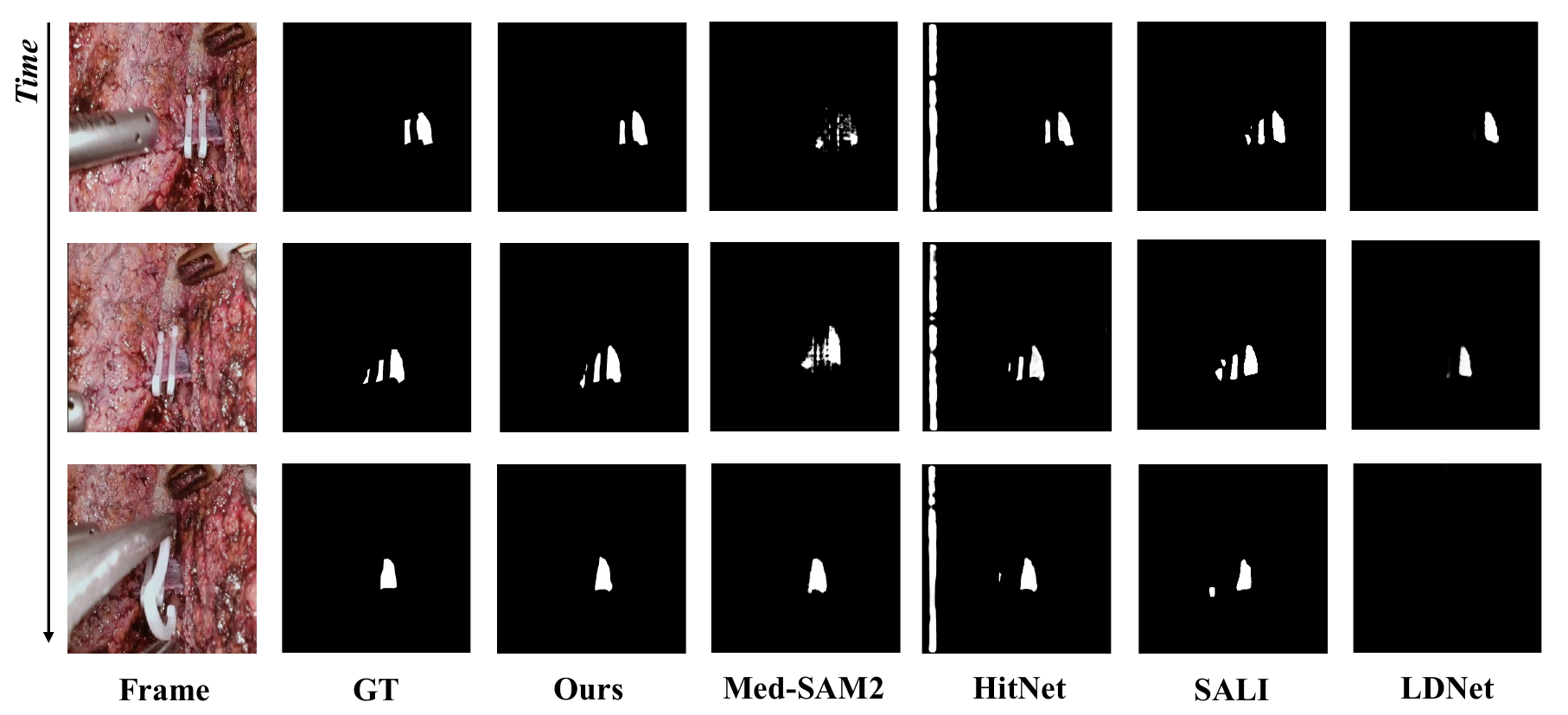}
    \caption{Visualization results of different methods on a challenge clip.}
    \label{fig: visible}
\end{figure}

As shown in Table \ref{tab:comparison_hepa}, our proposed HRVVS achieves state-of-the-art performance on the Hepa-SEG dataset, outperforming all baselines across most metrics.
Specifically, compared to the best-performing baseline, HRVVS achieves a relative improvement of +3.16\% in Jaccard index, +1.68\% in Dice coefficient, +4.20\% in F-measure, and +3.60\% in Enhanced-alignment measure. 
The only exception is the S-measure, where LDNet achieves a slightly higher score (0.8355 vs. 0.7878). However, LDNet exhibits a significantly lower Dice coefficient (0.2929), indicating that while it maintains high local consistency, it struggles to segment the complete target region (see Fig. \ref{fig: visible}).
Additionally, methods such as MemSAM and LDNet, which are optimized for ultrasound image segmentation, perform poorly on Hepa-SEG. This highlights the unique challenges posed by our dataset, where both spatial continuity and fine-grained vessel structures are critical for accurate segmentation.
\subsubsection{Qualitative Comparison.}

In Fig.~\ref{fig: visible}, we visualize the segmentation results of HRVVS alongside state-of-the-art methods on Hepa-SEG. HRVVS effectively captures fine details of hepatic vasculature, demonstrating superior segmentation accuracy and robustness in complex surgical scenes.

\subsection{Ablation Experiments}


We conduct the ablation study to evaluate the effectiveness of three main modules (i.e., the VAR branch, the MSIM module, and the DWFM module), and report the results in Tab~\ref{tab:ablation}.

\begin{table}[h]
  \caption{Ablation study on Hepa-SEG dataset. "VAR" denotes the VAR branch, MSIM and DWFM are two modules introduced above.}
  \centering
  \begin{adjustbox}{width=\textwidth}
  \begin{tabular}{c|c c c|c|c|c|c|c}
  \hline
  \textbf{Design} & \textbf{VAR} & \textbf{MSIM} & \textbf{DWFM} & \textbf{Jaccard \boldmath{$\uparrow$}} & \textbf{Dice \boldmath{$\uparrow$}} & \boldmath{${S}_\alpha \uparrow$} & \boldmath{$F_\beta^\omega \uparrow$} & \boldmath{$E_\phi^{mn} \uparrow$} \\
  \hline
  \hline
  basic & - & - & - & 0.4938 & 0.6122 & 0.7515 & 0.6311 & 0.8189 \\
  M1 & \checkmark & \checkmark & - & 0.4994 & 0.6233 & 0.7603 & 0.6307 & 0.8222 \\
  M2 & \checkmark & - & \checkmark & 0.5332 & 0.6442 & 0.7771 & 0.6712 & 0.8615 \\
  M3 & - & \checkmark & \checkmark & 0.5242 & 0.6384 & 0.7757 & 0.6613 & 0.8619 \\
  \hline
  \hline
  Ours & \checkmark & \checkmark & \checkmark & \textbf{0.5405} & \textbf{0.6532} & \textbf{0.7878} & \textbf{0.6769} & \textbf{0.8711} \\
  \hline
  \end{tabular}
  \label{tab:ablation}
  \end{adjustbox}
\end{table}

In this ablation study, we evaluate the impact of key components in our model on the Hepa-SEG dataset, specifically the VAR branch, Multi-scale Integration Module (MSIM), and Dynamic Weighted Feature Module (DWFM).

The baseline model, which excludes all three components, achieves a Jaccard index of 0.4938, a Dice coefficient of 0.6122, an $S_\alpha$ score of 0.7515, an $F_\beta^\omega$ score of 0.6311, and an $E_\phi^{mn}$ score of 0.8189. Adding only the VAR branch and MSIM (Model M1) slightly improves performance (Jaccard: 0.4994, Dice: 0.6233), suggesting their individual contributions are modest.

Incorporating VAR with DWFM (Model M2) leads to more substantial improvements (Jaccard: 0.5332, Dice: 0.6442), emphasizing DWFM’s effectiveness in feature refinement. Similarly, using MSIM and DWFM together (Model M3) enhances performance, though slightly less than M2. Finally, integrating all three components in the full model achieves the highest performance (Jaccard: 0.5405, Dice: 0.6532), demonstrating their complementary roles in improving segmentation accuracy.


\section{Conclusion}
This paper presents the first hepatic vasculature segmentation dataset under surgical video scenes, and a matching method based on hierarchical autoregressive residual priors. To address challenges in high-resolution surgical hepatectomy videos, our method proposes a VAR branch and a dynamic memory mechanism to embed them into a multi-view segmentation network. Experiments demonstrate that our HRVVS is capable of state-of-the-art results on Hepa-SEG and can be a critical baseline for video vasculature segmentation.

\begin{credits}
\subsubsection{\ackname}
 This work is supported by the Guangdong Science and Technology Department (No. 2024ZDZX2004) and Guangdong Provincial Key Lab of Integrated Communication, Sensing and Computation for Ubiquitous Internet of Things (No.2023B1212010007).

\subsubsection{\discintname}
The authors declare that they have no competing interests.
\end{credits}
%
%
%
\bibliographystyle{splncs04}
\bibliography{egbib}
%




\end{document}